%% file: AnonymousSubmission2027.tex
\documentclass[letterpaper]{article} 
\usepackage[preprint]{aaai2027}  
\usepackage[hyphens]{url}  
\usepackage{graphicx} 
\usepackage{natbib}  
\usepackage{caption} 
\usepackage{algorithm}
\usepackage{algorithmic}
\usepackage{amsmath}
\usepackage{amssymb}
\usepackage{multirow}
\usepackage{newfloat}
\usepackage{listings}
\DeclareCaptionStyle{ruled}{labelfont=normalfont,labelsep=colon,strut=off} 
\floatstyle{ruled}
\newfloat{listing}{tb}{lst}{}
\floatname{listing}{Listing}

\usepackage{booktabs}
\usepackage{array}
\newcommand{\ours}{\textsc{DreamGuard}}

\title{DreamGuard: Efficient Runtime Guardrail for LLM Agents via \\Risk-Aware World Model}
\author{
Wenhao Lin\textsuperscript{\rm 1},
Chenyu Yu\textsuperscript{\rm 1},
Xingwei Lin\textsuperscript{\rm 1}\corresponding,
Sicong Cao\textsuperscript{\rm 2}\corresponding,
Xiang Chen\textsuperscript{\rm 1},\\
Lei Xue\textsuperscript{\rm 3},
Le Yu\textsuperscript{\rm 2},
Letian Sha\textsuperscript{\rm 2},
Chunming Wu\textsuperscript{\rm 1}
}
\affiliations{
\textsuperscript{\rm 1}Zhejiang University\\
\textsuperscript{\rm 2}Nanjing University of Posts and Telecommunications\\
\textsuperscript{\rm 3}Sun Yat-sen University\\
\{22560233, 22621221, xwlin.roy, wuchunming\}@zju.edu.cn,
wasdnsxchen@gmail.com,\\
\{sicong.cao, yulele08, ltsha\}@njupt.edu.cn,
xuelei3@mail.sysu.edu.cn
}

\begin{document}

\maketitle

\begin{abstract}

As large language model (LLM) agents increasingly invoke external tools and interact with real-world systems, unsafe actions may cause irreversible consequences on external states, user data, and downstream services. 
Recent runtime guardrails mitigate such risks by checking proposed actions before execution, but many remain reactive: they primarily assess the apparent safety of the current action, lacking an explicit model of how risk evolves across the trajectory. This limitation creates a critical blind spot for long-horizon risks, where individually benign-looking actions can gradually drift the agent toward hazardous states.
In response, we propose \ours{}, a proactive guardrail for LLM agents built around a risk-aware world model. The world model maintains a compact recurrent latent state over the trajectory and predicts future latent states from which \ours{} derives immediate-hazard and prefix-risk evidence. It then fuses these multi-horizon signals into intervention decisions before execution.
Experiments across four benchmarks and an online guardrail evaluation show that \ours{} outperforms generic, reactive, and proactive guardrail baselines, achieves the best safety-utility trade-off among evaluated guardrails, and maintains an average end-to-end latency of 25 ms per call.
\end{abstract}


\section{Introduction}

Large language model (LLM) agents can now autonomously perform complex tasks in open-ended environments by invoking external tools and interacting with real-world systems~\cite{xi2025rise,qin2024toolllm,wang2024autonomousagents}. As these actions may affect external states, user data, and downstream services, ensuring agent runtime safety has become a critical concern~\cite{ma2025safetyatscale,yu2025trustworthyagents,gan2026navigating,deng2025agentsunderthreat}.

Recent research has introduced runtime guardrails for LLM agents,
which assess each proposed action in its trajectory context and
apply interventions to unsafe ones before
execution~\cite{xiang2025guardagent,chen2025shieldagent,wen2026policyguard}.
These guardrails can intercept immediate hazards, which would directly cause a hazardous outcome if executed at the current step, such as deleting files, leaking credentials, or invoking an unauthorized external API.

However, many existing runtime guardrails remain reactive: they assess the apparent safety of the current proposed action, without modeling how risk evolves over the trajectory or how the action may change future risk.
This limitation creates a critical blind spot for long-horizon risks, where early steps may appear benign in isolation, but gradually accumulate toward a hazardous outcome.
For example, an agent may locate an internal file, move it into a workspace folder, and later create a public sharing link; each operation can look routine locally, but the accumulated trajectory turns the final action into confidential data exposure. A guardrail that primarily assesses each proposed action based on its local state struggles to detect such risk accumulation.

In response, recent work has shifted toward proactive guardrails, which incorporate trajectory context or look-ahead predictions to assess how current actions may affect future risk. Yet, they still face several challenges in real-world deployments.
\textbf{(i)~Runtime guardrails must be efficient and lightweight.}
Many proactive guardrails rely on LLMs to process trajectory histories~\cite{li2026traces,chen2026safepred}. As the agent trajectory grows, these designs become constrained by context window length and require increasingly expensive autoregressive inference, often leading to multi-second latency per step. This strains the latency budget of real-time agent execution.
\textbf{(ii)~Prediction should preserve risk evidence.}
Proactive guardrails may predict future states or consequences and then assess them with separate safety rules or judges~\cite{liu2026safeagent,wang2025robosafe}. However, this separation can make the risk assessment depend on an incomplete prediction: when the prediction is not learned for risk discrimination, the safety assessment may lack the evidence needed to identify underlying risks. 
This calls for risk-aware latent states that expose hazardous transitions, rather than merely predicting generic future outcomes.
\textbf{(iii)~Runtime intervention requires calibrated multi-horizon risk.}
Proactive guardrails increasingly model risk evolution through trajectory memory, state tracking, or future-risk estimation~\cite{li2026traces,dhodapkar2026safetydrift,wang2026mage}. However, long-horizon evidence is often weak before hazards become explicit, while immediate-hazard evidence requires decisive intervention at the current action boundary. The challenge is to calibrate these complementary signals into an intervention rule that warns early without triggering excessive false alarms.


To address these challenges, we propose \ours{}, an efficient runtime guardrail for LLM agents built around a risk-aware world model. Specifically, to overcome the efficiency challenge (i), \ours{} maintains a fixed-dimensional state through GRU-based recurrent dynamics, bypassing the need to repeatedly process growing trajectories via LLMs. To tackle the risk evidence challenge (ii), \ours{} learns latent dynamics designed to explicitly preserve evidence of hazardous transitions in the predicted latent states. Finally, to address the multi-horizon risk intervention challenge (iii), \ours{} integrates both immediate-hazard and prefix-risk signals into a robust decision mechanism.
We evaluate \ours{} on four agent safety benchmarks covering both immediate hazards and long-horizon risks. Experiments show that \ours{} outperforms generic, reactive, and proactive guardrail baselines, while reducing average end-to-end latency to 25 ms per call. It further intervenes before the first hazard action in 96.3\% of unsafe long-horizon trajectories. In our online guardrail evaluation, \ours{} achieves the best safety--utility trade-off, reaching a 72.92\% safety rate while preserving 90.38\% task utility.

Our contributions are summarized as follows:
\begin{itemize}
    \item We introduce \ours{}, an efficient runtime guardrail built around a lightweight risk-aware world model that maintains a compact recurrent state and estimates multi-horizon risk before action execution. This design enables proactive intervention against both immediate hazards and long-horizon risks during real-time agent execution.

    \item We conduct extensive evaluations on four agent safety benchmarks. Experiments show that \ours{} outperforms guardrail baselines in mitigating multi-horizon risks, achieves earlier interventions against long-horizon risks, and maintains highly efficient end-to-end latency.

    \item We further validate \ours{} in an online guardrail evaluation, where it achieves the best safety--utility trade-off among evaluated guardrails, demonstrating its practical effectiveness during real-time agent execution.

\end{itemize}

\section{Related Work}

\paragraph{Guardrails for LLM Agents.} Generic LLM guardrails such as LlamaGuard~\cite{inan2023llamaguard} and WildGuard~\cite{han2024wildguard} focus on content moderation, classifying prompts or responses against harmful-content taxonomies. LLM agents fundamentally alter this paradigm: their outputs become actions that affect external environments, shifting the objective of guardrails from content moderation to action safety.
For instance, GuardAgent~\cite{xiang2025guardagent} and AGrail~\cite{luo2025agrail} translate task-specific safety requirements into executable action checks, while ShieldAgent~\cite{chen2025shieldagent} verifies trajectories against policy-derived temporal rules. Similarly, TS-Guard~\cite{mou2026toolsafe} and AgentDoG~\cite{liu2026agentdog} train specialized guard models for verifying action and trajectory safety. However, these methods predominantly operate as reactive guardrails, failing to effectively identify long-horizon risks that emerge over multi-step agent interactions.

To mitigate long-horizon risks, a further line of work is moving toward proactive guardrailing by incorporating trajectory context or look-ahead prediction. For instance, SafetyDrift~\cite{dhodapkar2026safetydrift} and ProbGuard~\cite{wang2025probguard} estimate future violation probabilities over hand-crafted discrete or symbolic safety states, while MAGE~\cite{wang2026mage} and SafeAgent~\cite{liu2026safeagent} maintain safety-relevant memory or runtime state for risk assessment. Recent world model-based guardrails such as SafePred~\cite{chen2026safepred} and SafeMCP~\cite{wang2026safemcp} further use LLM look-ahead reasoning for future risk prediction. Despite these advances, these methods still struggle to provide efficient runtime protection against multi-horizon risks, often relying on LLM-based reasoning or domain-specific action spaces.

\paragraph{World Models for Agent Safety.} 
For proactive agent guardrailing, anticipating the consequences of proposed actions is essential to intervene before unsafe outcomes occur. World models have long been used in model-based decision making to capture environment dynamics and support planning over predicted futures. Latent world models such as PlaNet~\cite{hafner2019planet}, DreamerV3~\cite{hafner2025dreamerv3}, and TD-MPC2~\cite{hansen2024tdmpc2} learn compact latent dynamics from interaction data and use predicted futures for planning or control. This idea has also been adapted to LLM agents: WebDreamer~\cite{gu2024webdreamer} and WMA~\cite{chae2025wma} simulate action outcomes for web navigation. Beyond task performance, SafePred~\cite{chen2026safepred} uses an LLM-based world model to predict action consequences and feed risk estimates into a decision loop, and SafeMCP~\cite{wang2026safemcp} relies on LLM-based look-ahead reasoning over environment dynamics. Despite these advances, existing world models for LLM agent safety often rely on LLM-based look-ahead reasoning, which can incur high per-action cost. \ours{} instead uses lightweight GRU-based recurrent dynamics to maintain a compact risk state from trajectory data, enabling efficient risk estimation for runtime intervention.

\section{Methodology}

\begin{figure*}[t]
\centering
\includegraphics[width=0.95\textwidth]{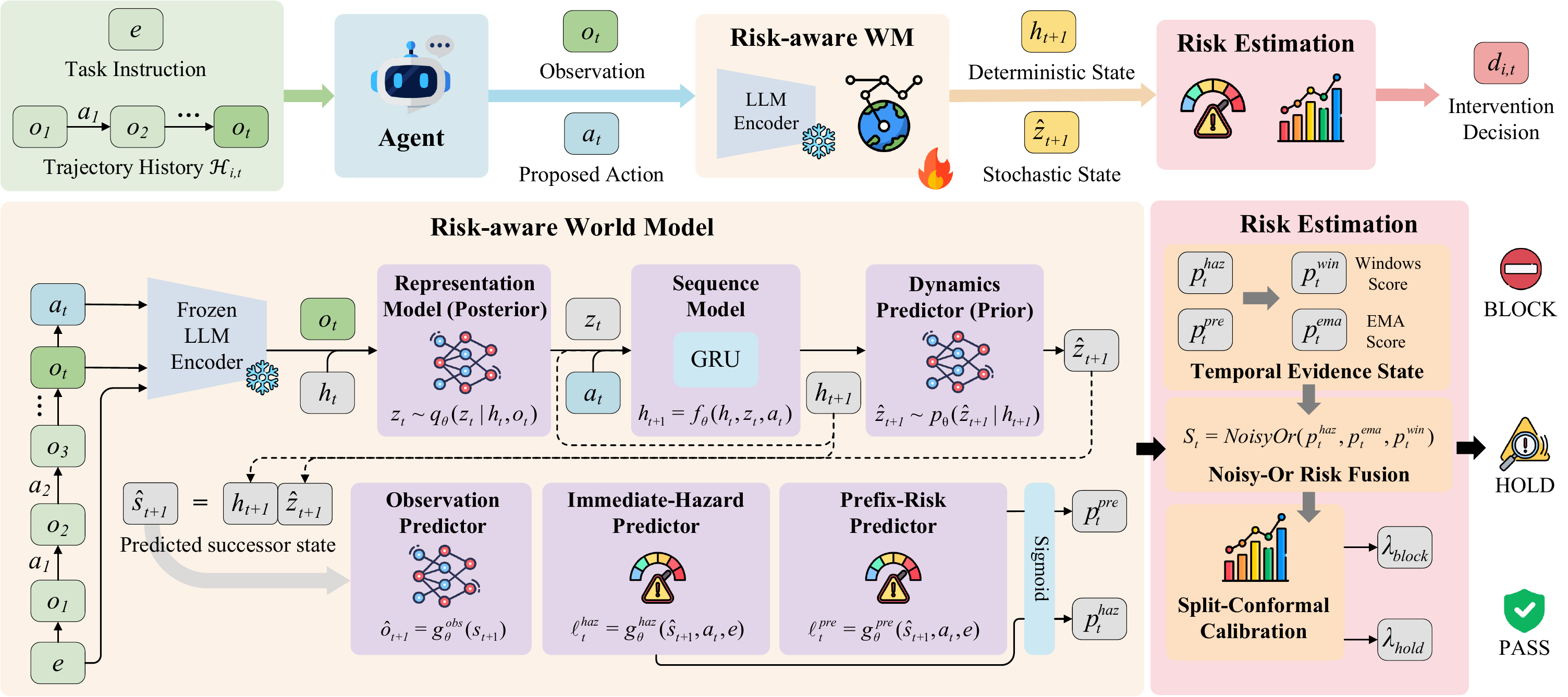}
\caption{Overview of \ours{}. 
(1) Risk-aware World Model: encodes the trajectory prefix and proposed action into a recurrent latent state, predicts the successor latent state, and produces immediate-hazard and prefix-risk scores. 
(2) Multi-horizon Risk Estimation: aggregates temporal risk evidence and applies a calibrated intervention rule to output a runtime decision before action execution.}
\label{fig:overview}
\end{figure*}

\subsection{Problem Formulation}

We consider an LLM agent that interacts with an external
environment over a trajectory $i$ of length $T_i$. Runtime risks may
arise from unsafe user intent, adversarial attack, or imperfect agent reasoning. These risks can be immediate,
where an action is hazardous at the current step, or
long-horizon, where an action may appear benign locally but contributes
to a hazardous outcome through the accumulated trajectory context. We
formulate guardrailing as a step-level decision problem. At
each step $t$, before the proposed action is executed, the guardrail $\mathcal{G}$
receives
\begin{equation}
x_{i,t}=(e,\mathcal{H}_{i,t},a_{i,t}),
\end{equation}
where $e$ denotes the task instruction given to the agent, $a_{i,t}$ denotes the agent’s proposed action at step $t$, and $o_{i,t}$ denotes the current environment observation available to the agent before it executes $a_{i,t}$. The trajectory prefix
$
\mathcal{H}_{i,t}=(o_{i,1},a_{i,1},o_{i,2},\ldots,a_{i,t-1},o_{i,t})
$
contains the interaction history available up to $a_{i,t}$.
The guardrail outputs a step-level decision:
\begin{equation}
d_{i,t}=\mathcal{G}(x_{i,t})\in\{\mathrm{PASS},\mathrm{HOLD},\mathrm{BLOCK}\}.
\end{equation}
$\mathrm{PASS}$ allows the action to execute, $\mathrm{HOLD}$ interrupts execution when the evidence is low-confidence or reflects accumulated risk and triggers a conservative intervention, and $\mathrm{BLOCK}$ interrupts execution and prevents the action.

For a hazard trajectory, let $t_i^{\mathrm{haz}}$ denote the first hazard step, the earliest step whose proposed action would trigger a concrete hazardous outcome or transition if executed. The primary runtime objective is to estimate step-level risk and intervene before unsafe actions are executed, returning either $\mathrm{HOLD}$ or $\mathrm{BLOCK}$ at $t_i^{\mathrm{haz}}$ while minimizing false alerts on safe trajectories.

\subsection{Overview of \ours{}}

The goal of \ours{} is to equip runtime guardrails with proactive risk prediction, enabling them to identify proposed actions that may cause hazardous outcomes. As shown in Figure~\ref{fig:overview}, \ours{} consists of two modules. The risk-aware world model maintains a compact latent state over the trajectory, and estimates immediate-hazard and prefix-risk evidence from the predicted latent state. The multi-horizon risk estimation module then aggregates temporal evidence and maps the fused score to calibrated intervention decisions. Algorithm~\ref{alg:dreamguard} formalizes the runtime workflow.

\subsection{Risk-Aware World Model}

\paragraph{Recurrent World Model Architecture.}
Learning a generative model by reconstructing raw future observations
provides a rich signal, but accurate raw reconstruction is costly and
does not directly determine whether an agent action is safe. For
runtime guardrailing, the model must instead capture how the
trajectory prefix evolves under proposed actions and preserve the
safety-relevant information needed for intervention. Inspired by DreamerV3~\cite{hafner2025dreamerv3} and
TD-MPC2~\cite{hansen2024tdmpc2}, \ours{} builds a recurrent
state-space world model (RSSM) with implicit,
risk-aware latent dynamics for predicting successor states and estimating risk before action execution.

The \ours{} world model architecture is shown in Figure~\ref{fig:overview} and consists of six components:
\begin{equation}
\begin{alignedat}{2}
&\text{Sequence model:}        &&\quad h_t = f_\theta(h_{t-1},z_{t-1},a_{t-1}), \\
&\text{Representation model:}  &&\quad z_t \sim q_\theta(z_t\mid h_t,o_t), \\
&\text{Dynamics predictor:}    &&\quad \hat{z}_t \sim p_\theta(\hat{z}_t\mid h_t), \\
&\text{Observation predictor:} &&\quad \hat{o}_{t+1} = g^{\mathrm{obs}}_\theta(s_{t+1}), \\
&\text{Immediate-hazard predictor:}           &&\quad \ell^{\mathrm{haz}}_t = g^{\mathrm{haz}}_\theta(\hat{s}_{t+1},a_t,e), \\
&\text{Prefix-risk predictor:}      &&\quad \ell^{\mathrm{pre}}_t = g^{\mathrm{pre}}_\theta(\hat{s}_{t+1},a_t,e).
\end{alignedat}
\end{equation}
Here, $\sim$ marks stochastic RSSM variables drawn from posterior or prior distributions, whereas $=$ marks deterministic neural readouts. At step $t$, the representation model conditions on the current observation to infer $z_t$; together with the recurrent state $h_t$, which summarizes the trajectory prefix, this gives the posterior model state $s_t=(h_t,z_t)$. The proposed action $a_t$ is then fed to the sequence model to obtain $h_{t+1}$, from which the dynamics predictor produces the successor prior $\hat z_{t+1}$. The resulting prior state $\hat s_{t+1}=(h_{t+1},\hat z_{t+1})$ supports immediate-hazard and prefix-risk estimation before execution. During training, the observation predictor also uses the successor posterior state $s_{t+1}=(h_{t+1},z_{t+1})$, providing an auxiliary prediction objective for shaping the latent dynamics.

\ours{} uses two training stages: world model pretraining and
risk-supervised training. The first stage learns recurrent latent
dynamics from trajectory data, while the second stage shapes the
learned state with immediate-hazard and prefix-risk supervision.

\paragraph{World Model Pretraining.}
Given a trajectory sequence sampled from the world model
training dataset $\mathcal{D}_{\mathrm{wm}}$, \ours{} optimizes the world model
parameters $\theta$ with the following objective. Let $m^{\mathrm{wm}}_t\in\{0,1\}$ denote the pretraining mask at step $t$. The objective is
\begin{equation}
\begin{aligned}
&\mathcal{L}_{\mathrm{WM}}(\theta)
=
\mathbb{E}_{\tau\sim\mathcal{D}_{\mathrm{wm}},\,z\sim q_\theta}
\Bigg[
\frac{1}{\sum_{t=1}^{T}m^{\mathrm{wm}}_t}
\sum_{t=1}^{T}m^{\mathrm{wm}}_t \\
&\qquad\times
\big(
\lambda_{\mathrm{pred}}\mathcal{L}^{t}_{\mathrm{pred}}
+\lambda_{\mathrm{dyn}}\mathcal{L}^{t}_{\mathrm{dyn}}
+\lambda_{\mathrm{rep}}\mathcal{L}^{t}_{\mathrm{rep}}
\big)
\Bigg].
\end{aligned}
\end{equation}
The prediction loss $\mathcal{L}^{t}_{\mathrm{pred}}$ operates in embedding space, training the observation predictor to predict the embedding of the observation after action execution rather than the raw observation itself:
\begin{equation}
\mathcal{L}^{t}_{\mathrm{pred}}
=
1-\cos(\bar{\hat{o}}_{t+1},\bar{o}_{t+1}),
\end{equation}
where $\bar{\hat{o}}_{t+1}$ and $\bar{o}_{t+1}$ denote batch-centered
prediction and target embeddings. The dynamics and representation
losses regularize the stochastic state with opposite stop-gradient
directions:
\begin{equation}
\begin{aligned}
\mathcal{L}^{t}_{\mathrm{dyn}}
&=
\max\!\Big(
\mathrm{KL}\!\big[
\mathrm{sg}(q_\theta(z_t\mid h_t,o_t))
\,\|\, p_\theta(\hat{z}_t\mid h_t)
\big],
\kappa
\Big), \\
\mathcal{L}^{t}_{\mathrm{rep}}
&=
\max\!\Big(
\mathrm{KL}\!\big[
q_\theta(z_t\mid h_t,o_t)
\,\|\, \mathrm{sg}(p_\theta(\hat{z}_t\mid h_t))
\big],
\kappa
\Big).
\end{aligned}
\end{equation}
where $\mathrm{sg}(\cdot)$ denotes stop-gradient and $\kappa$ is the
free-bits threshold. The dynamics loss $\mathcal{L}^{t}_{\mathrm{dyn}}$ trains the prior to predict posterior states, while the representation loss $\mathcal{L}^{t}_{\mathrm{rep}}$ encourages posterior
states to remain predictable rather than encoding arbitrary
observation details.

\paragraph{Risk-Supervised Training.}
After world model pretraining, \ours{} has learned recurrent latent dynamics from trajectory data. Risk-supervised training then shapes the predicted successor states with two safety targets. Let $y^{\mathrm{haz}}_{i,t}$ denote the hazardous label for trajectory $i$ at step $t$. 

For prefix-risk supervision, we assign a decayed positive target to precursor steps within horizon $K$ and a target value of 1 to immediately hazardous steps:
\begin{equation}
y^{\mathrm{pre}}_{i,t}
=
\begin{cases}
1, & y^{\mathrm{haz}}_{i,t}=1,\\
\exp\left(-({t_i^{\mathrm{haz}}-t-1})/\rho\right),
& 1\le t_i^{\mathrm{haz}}-t\le K,\\
0, & \text{otherwise},
\end{cases}
\end{equation}
where decay temperature $\rho$ controls how quickly the soft prefix label decays with distance from $t_i^{\mathrm{haz}}$. Trajectories without a hazard have $y^{\mathrm{pre}}_{i,t}=0$ for all steps.


These two safety targets are applied only on the supervised prefix of each trajectory. Let $m^{\mathrm{safe}}_t$ denote the resulting mask over supervised steps. The safety objective averages the immediate-hazard and prefix-risk losses over this supervised prefix:
\begin{equation}
\begin{aligned}
&\mathcal{L}_{\mathrm{safety}}(\theta)
=
\mathbb{E}_{\tau\sim\mathcal{D}_{\mathrm{safety}},\,
z\sim q_\theta,\,\hat{z}\sim p_\theta}
\Bigg[
\frac{1}{\sum_{t=1}^{T}m^{\mathrm{safe}}_t} \\
&\qquad\times
\sum_{t=1}^{T}m^{\mathrm{safe}}_t
\big(
\mathcal{L}^{t}_{\mathrm{haz}}
+\mathcal{L}^{t}_{\mathrm{pre}}
\big)
\Bigg].
\end{aligned}
\end{equation}
The two losses supervise complementary signals: immediate-hazard risk for actions that would trigger a hazardous outcome at the current step, and prefix-risk for trajectory prefixes indicating accumulating long-horizon risk:
\begin{align}
\mathcal{L}^{t}_{\mathrm{haz}}
&=
\operatorname{BCE}_{\mathrm{logit}}(\ell^{\mathrm{haz}}_{i,t},y^{\mathrm{haz}}_{i,t}), \\
\mathcal{L}^{t}_{\mathrm{pre}}
&=
\operatorname{BCE}_{\mathrm{logit}}(\ell^{\mathrm{pre}}_{i,t},y^{\mathrm{pre}}_{i,t}).
\end{align}
These safety losses update the partial unfrozen world model modules on the computational path to the immediate-hazard and prefix-risk predictors, while keeping the GRU transition core fixed. This partial unfreezing lets multi-horizon risk supervision shape the latent state, preserving information predictive of hazardous outcomes and risk accumulation. As a result, the latent dynamics become risk-aware rather than merely predictive of generic next states.

\paragraph{Architecture details.}
The sequence model is implemented as a GRU-based transition with LayerNorm on the concatenated stochastic state and action input. The remaining projection and predictor modules are MLPs with Linear layers, LayerNorm, and SiLU activations. The stochastic state is represented as factorized categorical variables and sampled with straight-through gradients.

\subsection{Multi-Horizon Risk Estimation}

At runtime, \ours{} uses the world model to predict the successor prior state for the proposed action and maps it to two complementary risk scores: the immediate-hazard score $p^{\mathrm{haz}}_t$ for actions that may directly trigger a hazardous outcome, and the prefix-risk score $p^{\mathrm{pre}}_t$ for accumulating long-horizon risk. Because prefix-risk scores are designed to be sensitive to weak precursor evidence, DreamGuard smooths them with a temporal evidence state $B_t=(p^{\mathrm{ema}}_t,p^{\mathrm{win}}_t)$:
\begin{align}
{p}^{\mathrm{ema}}_t
&=
\beta {p}^{\mathrm{ema}}_{t-1}
+(1-\beta)p^{\mathrm{pre}}_t, \\
{p}^{\mathrm{win}}_t
&=
\frac{1}{|\mathcal{W}_t|}
\sum_{j\in\mathcal{W}_t}p^{\mathrm{pre}}_j,
\end{align}
where $\beta\in[0,1)$ controls how strongly past prefix-risk evidence is retained, and $\mathcal{W}_t$ denotes the most recent window of at most $W$ prefix-risk scores. The exponential-moving-average score $p^{\mathrm{ema}}_t$ captures weak risk evidence that persists over time, while the recent-window score $p^{\mathrm{win}}_t$ preserves transient increases in prefix-risk evidence without letting older evidence dominate.
\ours{} then aggregates immediate-hazard and prefix-risk evidence into a fused runtime risk score $S_t$ using a bounded noisy-or rule:
\begin{equation}
S_t
=
\operatorname{NoisyOr}\!
\left(
p^{\mathrm{haz}}_t,{p}^{\mathrm{ema}}_t,{p}^{\mathrm{win}}_t
\right),
\end{equation}
Here, $\operatorname{NoisyOr}(\cdot)$ applies calibrated fusion weights and bounds each weighted term to $[0,1]$ before multiplication, so $S_t$ increases when any evidence channel is strong while remaining in $[0,1]$. The calibrated intervention rule $\mathcal{I}$ maps $(p^{\mathrm{haz}}_t,S_t)$ to an intervention decision before execution:
\begin{equation}
\mathcal{I}(p^{\mathrm{haz}}_t,S_t)=
\begin{cases}
\mathrm{BLOCK}, & p^{\mathrm{haz}}_t \ge \lambda_{\mathrm{block}},\\
\mathrm{HOLD}, & S_t \ge \lambda_{\mathrm{hold}},\\
\mathrm{PASS}, & \text{otherwise.}
\end{cases}
\end{equation}
Both thresholds are calibrated on safe trajectories using split-conformal quantiles. 
The $\mathrm{BLOCK}$ threshold $\lambda_{\mathrm{block}}$ is calibrated from $p^{\mathrm{haz}}_t$ with a stricter false-alert budget, while the $\mathrm{HOLD}$ threshold $\lambda_{\mathrm{hold}}$ is calibrated from $S_t$. 
For $\lambda_{\mathrm{hold}}$, let
$M_i=\max_{1\le t\le T_i}S_{i,t}$ denote the maximum fused score along each calibration trajectory $i\in\mathcal{D}^{\mathrm{safe}}_{\mathrm{cal}}$. 
Sorting $M_1,\ldots,M_n$ as $M_{(1)}\le\cdots\le M_{(n)}$, \ours{} sets
$
\lambda_{\mathrm{hold}}
=
M_{(k)}$
and
$ k=\min\!\left(\left\lceil (n+1)(1-\alpha)\right\rceil,n\right),
$
where $n=|\mathcal{D}^{\mathrm{safe}}_{\mathrm{cal}}|$ and $\alpha$ is the target false-alert level on safe trajectories. 
Because $M_i$ is the maximum score over all steps in trajectory $i$, this threshold calibrates trajectory-level false alerts: a safe trajectory is counted as alerted if any step triggers $\mathrm{HOLD}$. 
The same construction is applied to immediate-hazard scores $p^{\mathrm{haz}}_t$ to obtain $\lambda_{\mathrm{block}}$. 

\begin{algorithm}[tb]
\captionsetup{labelfont=bf}
\caption{Workflow of \ours{}}
\label{alg:dreamguard}
\renewcommand{\algorithmiccomment}[1]{\hfill\textcolor{blue}{$\triangleright$~#1}}
\textbf{Input}: Guardrail $\mathcal{G}=(\theta,\mathcal{I})$;
task-instruction representation $e$;
online pre-action stream $\{(o_t,a_t)\}_{t=1}^{T}$\\
\textbf{Output}: Decision log $D$
\begin{algorithmic}[1]
\STATE $h_1 \leftarrow \mathrm{Init}_\theta(e)$;\;
$B_0 \leftarrow (0,\emptyset)$;\;
$D \leftarrow \emptyset$
\COMMENT{decision log}
\FOR{$t = 1,\ldots,T$}
\STATE \textbf{// Phase 1: successor-state prediction}
\STATE $z_t \sim q_\theta(z_t \mid h_t, o_t)$
\COMMENT{infer current posterior state}
\STATE $h_{t+1} \leftarrow f_\theta(h_t, z_t, a_t)$
\COMMENT{imagine action effect}
\STATE $\hat{z}_{t+1} \sim p_\theta(\hat{z}_{t+1} \mid h_{t+1})$
\COMMENT{predict successor prior}
\STATE $\hat{s}_{t+1} \leftarrow (h_{t+1}, \hat{z}_{t+1})$
\STATE \textbf{// Phase 2: multi-horizon risk estimation}
\STATE $p^{\mathrm{haz}}_t \leftarrow
\sigma\!\big(g^{\mathrm{haz}}_\theta(\hat{s}_{t+1}, a_t, e)\big)$
\STATE $p^{\mathrm{pre}}_t \leftarrow
\sigma\!\big(g^{\mathrm{pre}}_\theta(\hat{s}_{t+1}, a_t, e)\big)$
\STATE $B_t \leftarrow \mathrm{Update}(B_{t-1}, p^{\mathrm{pre}}_t)$
\COMMENT{accumulate prefix-risk evidence}
\STATE $S_t \leftarrow \mathrm{Fuse}(p^{\mathrm{haz}}_t, B_t)$
\STATE \textbf{// Phase 3: calibrated pre-action intervention}
\STATE $d_t \leftarrow \mathcal{I}(p^{\mathrm{haz}}_t, S_t)$
\STATE $D \leftarrow D \cup \{(t, d_t)\}$
\IF{$d_t \neq \mathrm{PASS}$}
\STATE \textbf{return} $D$
\COMMENT{HOLD or BLOCK before execution}
\ENDIF
\ENDFOR
\STATE \textbf{return} $D$
\end{algorithmic}
\end{algorithm}

\section{Experiments}

\begin{table*}[t]
\centering
\small
\setlength{\tabcolsep}{1.5pt}
\begin{tabular*}{\linewidth}{@{\extracolsep{\fill}}l *{16}{c}@{}}
\toprule
\multirow{2}{*}{Method}
& \multicolumn{4}{c}{SafetyDrift}
& \multicolumn{4}{c}{AgentDojo}
& \multicolumn{4}{c}{ASB}
& \multicolumn{4}{c}{ASSE-Security} \\
\cmidrule(lr){2-5}\cmidrule(lr){6-9}
\cmidrule(lr){10-13}\cmidrule(lr){14-17}
& F1$\uparrow$ & SR$\uparrow$ & FPR$\downarrow$ & Lat.$\downarrow$
& F1$\uparrow$ & SR$\uparrow$ & FPR$\downarrow$ & Lat.$\downarrow$
& F1$\uparrow$ & SR$\uparrow$ & FPR$\downarrow$ & Lat.$\downarrow$
& F1$\uparrow$ & SR$\uparrow$ & FPR$\downarrow$ & Lat.$\downarrow$ \\
\midrule
\multicolumn{17}{l}{\textbf{Generic Guardrail}} \\
\midrule
Llama-Guard-4-12B
& 37.3 & 40.7 & 77.8 & 0.130
& \underline{78.3} & 67.4 & \underline{6.3} & 0.446
& 28.0 & 16.3 & \textbf{0.0} & 0.158
& 59.4 & 42.7 & \underline{1.2} & 0.156 \\
\midrule
\multicolumn{17}{l}{\textbf{Reactive Guardrails}} \\
\midrule
PolicyGuard
& 26.4 & 22.2 & 66.7 & \underline{0.063}
& 47.3 & 36.8 & 24.7 & 0.128
& \underline{70.4} & \underline{54.4} & 1.9 & 0.088
& \underline{79.1} & \textbf{85.1} & 32.5 & 0.076 \\
GuardAgent
& 74.0 & \underline{88.9} & 66.7 & 6.655
& 63.8 & 57.8 & 30.6 & 7.029
& 62.6 & 45.7 & 3.5 & 6.040
& 53.8 & 44.3 & 22.2 & 5.338 \\
AgentDoG-1.5
& 54.8 & 55.6 & 59.3 & 1.139
& \textbf{80.0} & \underline{76.6} & 19.6 & 1.182
& 40.4 & 25.3 & 1.9 & 1.122
& 53.6 & 37.0 & \textbf{0.9} & 0.845 \\
\midrule
\multicolumn{17}{l}{\textbf{Proactive Guardrails}} \\
\midrule
SafePred
& 45.3 & 37.0 & 44.4 & 7.686
& 48.8 & 33.5 & \textbf{5.1} & 8.012
& 54.5 & 37.5 & \underline{1.2} & 7.315
& 20.5 & 14.7 & 30.8 & 19.383 \\
TRACES
& \underline{87.1} & \textbf{96.3} & \underline{33.3} & 0.081
& 46.6 & 35.0 & 20.0 & \underline{0.106}
& 27.3 & 16.2 & 21.8 & \underline{0.073}
& 70.7 & 72.8 & 36.1 & \underline{0.066} \\
\ours{}
& \textbf{96.4} & \textbf{96.3} & \textbf{3.7} & \textbf{0.027}
& 74.9 & \textbf{76.9} & 29.4 & \textbf{0.034}
& \textbf{82.1} & \textbf{74.2} & 13.6 & \textbf{0.023}
& \textbf{82.9} & \underline{77.2} & 9.8 & \textbf{0.016} \\
\bottomrule
\end{tabular*}
\caption{Main runtime guardrail results across four agent safety
benchmarks. \textbf{F1}, \textbf{SR}, and \textbf{FPR} (in \%) evaluate trajectory-level hazardous trajectory detection; \textbf{Latency} reports end-to-end wall-clock seconds per guardrail call. Best results are boldfaced and second-best results
are underlined.}
\label{tab:main}
\end{table*}

\subsection{Experimental Setup}


\paragraph{Benchmarks.}
We evaluate \ours{} on four benchmarks covering both long-horizon risk accumulation and immediate hazards.
(1)~SafetyDrift~\cite{dhodapkar2026safetydrift} contains multi-step trajectories in which risk accumulates before a safety constraint is violated.
(2)~AgentDojo~\cite{debenedetti2024agentdojo} provides tool-use tasks with prompt injection and environment-mediated risks.
(3)~Agent Security Bench (ASB)~\cite{zhang2025asb} covers direct and indirect attacks against LLM agents.
(4)~ASSE-Security~\cite{luo2025agentauditor} covers data leakage, credential abuse, unauthorized access, and prompt injection.
\ours{} is trained and calibrated on SafetyDrift, using disjoint trajectory splits. SafetyDrift results are reported on the held-out test split, and the other benchmarks are evaluated with the same calibrated thresholds.

\paragraph{Step-Level Annotation.}
Most existing benchmarks provide only trajectory-level labels, leaving step-level hazard labels unspecified. To address this, we convert the four benchmarks into a common step-level trajectory format and annotate the hazard attribute of each step using benchmark-specific rules, with initial prelabeling from an LLM and independent review by two human annotators. The first hazard step is then identified as the earliest step labeled hazardous in each unsafe trajectory.

\paragraph{Baselines.}
We compare \ours{} with three groups of guardrails.
(1) Generic guardrail: Llama-Guard-4-12B~\cite{llamaguard4}, a broadly deployed
safety classifier adapted to agent trajectories.
(2) Reactive guardrails: PolicyGuard~\cite{wen2026policyguard},
GuardAgent~\cite{xiang2025guardagent}, and
AgentDoG-1.5~\cite{liu2026agentdog15}, which primarily assess the apparent
safety of the current proposed action or explicit policy constraints.
(3) Proactive guardrails: SafePred~\cite{chen2026safepred} and
TRACES~\cite{li2026traces}, which maintain persistent state, track
evolving risk, or predict future consequences to address long-horizon
risks.

\paragraph{Metrics.}
We evaluate guardrails at both trajectory and step levels.
(1)~Following prior runtime guardrail evaluations, we report safety
rate (\textbf{SR}), \textbf{F1}, false positive rate (\textbf{FPR}), and \textbf{Latency} on the full
trajectory. SR measures the fraction of unsafe trajectories in
which the guardrail triggers an intervention or unsafe decision.
Latency is the end-to-end wall-clock time per guardrail call.
(2)~Since trajectory-level metrics do not capture whether intervention
is timely, we further report Pre-Hazard Intervention Recall (\textbf{PHIR}),
the fraction of unsafe trajectories where the first intervention occurs
strictly before the first hazard step. We also report Mean Advance
Steps (\textbf{MAS}), the average number of steps by which successful
interventions precede the first hazard step. Formal metric definitions
are provided in Appendix~A.

\paragraph{Implementation Details.}
\ours{} uses Qwen3-4B-Instruct-2507 as a frozen embedding encoder. We train \ours{} on a single H100 GPU using SafetyDrift as the source benchmark. For a fair comparison, all trainable baselines use the same training data, and all baselines are evaluated with the same input fields and metric protocol whenever applicable. Additional implementation details are provided in Appendix~C.
\subsection{Main Results}
Table~\ref{tab:main} summarizes the main results across benchmarks covering immediate hazards and long-horizon risks. \ours{} shows the strongest overall F1 and SR performance across all benchmarks while keeping FPR low. It also achieves the lowest latency on all benchmarks.

SafetyDrift is designed to evaluate long-horizon risk. On this benchmark, \ours{} achieves the best F1 of 96.4\% and the highest SR of 96.3\%, while reducing FPR to 3.7\%. This suggests that \ours{} can preserve weak early risk evidence until it becomes actionable, rather than overreacting to benign prefixes. In particular, its recurrent latent state preserves trajectory context, allowing the prefix-risk predictor to accumulate weak precursor evidence into a stable score for earlier intervention.

AgentDojo, ASB, and ASSE-Security mainly focus on immediate hazards. On these benchmarks, DreamGuard achieves the best SR on AgentDojo and the best F1/SR on ASB, while obtaining the best F1 on ASSE-Security with lower FPR than PolicyGuard, the strongest SR baseline. This indicates that DreamGuard can identify immediate hazards while avoiding excessive false alerts. The contrast with SafetyDrift further highlights \ours{}'s consistent performance across risk horizons: TRACES reaches 87.1\% F1 on SafetyDrift but drops to 27.3\% on ASB, while PolicyGuard reaches 79.1\% F1 on ASSE-Security but only 26.4\% on SafetyDrift. \ours{} avoids this sharp trade-off by maintaining multi-horizon risk evidence.

\ours{} is the most low-latency guardrail in Table~\ref{tab:main}, with
an average end-to-end latency of 0.025 s per call. Compared
with specialized efficient guardrails, it is 3.3$\times$ faster than
TRACES and 3.6$\times$ faster than PolicyGuard. The gap is much larger
against LLM-reasoning guardrails: \ours{} is 250.6$\times$ faster than
GuardAgent and 424.0$\times$ faster than SafePred. This efficiency comes
from its fixed-dimensional recurrent state and lightweight predictors,
avoiding repeated full-trajectory processing or online LLM reasoning.

\subsection{Timing Analysis}

\begin{table}[t]
\centering
\small
\setlength{\tabcolsep}{1.5pt}
\begin{tabular}{@{}lcccc@{}}
\toprule
Method
& SafetyDrift
& AgentDojo
& ASB
& ASSE \\
\midrule
LG-4-12B
& 40.7 / 2.73 & 0.6 / 0.01 & 0.0 / 0.00 & 5.7 / 0.15 \\
\midrule
PolicyGuard
& 16.1 / 1.33 & 3.3 / \textbf{0.36}
& \underline{8.5} / \underline{0.17} & 6.0 / 0.09 \\
GuardAgent
& \underline{87.1} / \underline{3.07}
& \underline{13.8} / 0.27 & 3.2 / 0.08 & 4.6 / 0.17 \\
AgentDoG-1.5
& 35.5 / 2.00 & 1.8 / 0.03 & 1.3 / 0.07 & 2.4 / 0.07 \\
\midrule
SafePred
& 25.8 / 1.73 & 0.9 / 0.03 & 0.9 / 0.03 & 1.4 / 0.17 \\
TRACES
& \underline{87.1} / 2.00 & 5.7 / 0.29
& 0.4 / 0.07 & \textbf{48.4} / \textbf{0.78} \\
\ours{}
& \textbf{96.3} / \textbf{3.63}
& \textbf{16.8} / \underline{0.33}
& \textbf{17.3} / \textbf{0.27}
& \underline{34.8} / \underline{0.61} \\
\bottomrule
\end{tabular}
\caption{Step-level timing results across four benchmarks.
Each cell reports PHIR/MAS.
ASSE denotes ASSE-Security. LG-4-12B denotes Llama-Guard-4-12B.
Best results are boldfaced and
second-best results are underlined.}
\label{tab:timing}
\end{table}

We further evaluate \ours{} at the step level by measuring whether a
guardrail intervenes before the first hazard action. On SafetyDrift, \ours{}
achieves the highest PHIR (96.3\%) and MAS
(3.63), outperforming both reactive and proactive guardrails.
This gain is enabled by its recurrent risk-aware state and prefix-risk
evidence aggregation, which preserve weak precursor signals before the
hazard action becomes explicit. On the more immediate-hazard
benchmarks, \ours{} also obtains the highest PHIR on AgentDojo and
ASB, and the highest MAS on ASB, suggesting that its immediate-hazard
and prefix-risk signals remain useful near the action boundary.
Overall, these results show that \ours{} can anticipate hazardous
actions before execution and trigger earlier runtime intervention.

\subsection{Online Guardrail}

\begin{figure}[t]
\centering
\includegraphics[width=\columnwidth]{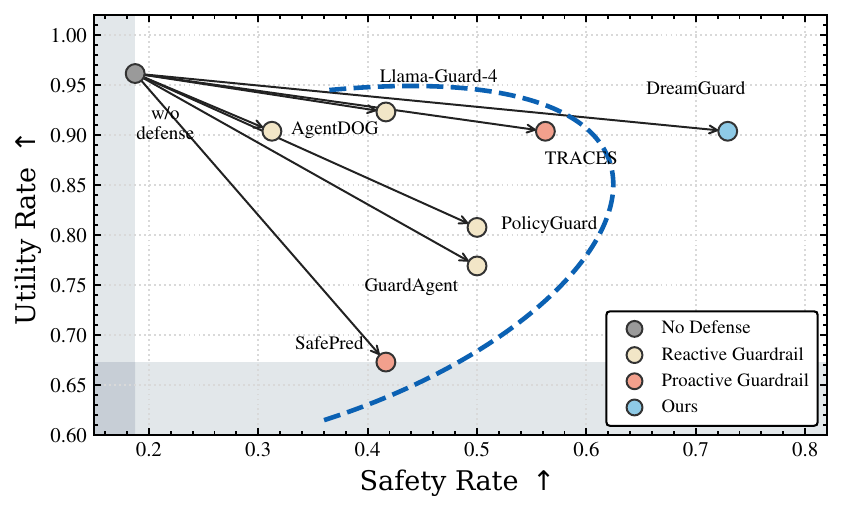}
\caption{Defense performance comparison in the online guardrail
setting. The blue dashed line represents the safety--utility frontier
across all baselines.}
\label{fig:online-pareto}
\end{figure}

We further evaluate \ours{} in providing online guardrails for tool-use agents. Specifically, we use GPT-5.1~\cite{openai2025gpt51systemcard} as the task agent and integrate each
guardrail as a runtime verification module alongside the agent. As shown in Figure~\ref{fig:online-pareto}, \ours{} occupies
a favorable region of the trade-off space: it achieves the highest
Safety Rate (72.92\%) while maintaining high Utility Rate
(90.38\%), placing it beyond the safety--utility frontier
formed by existing guardrails. This indicates that \ours{} improves
online unsafe-action interception without collapsing benign-task
utility.

\subsection{Ablation Studies}

To evaluate the contribution of each component, we conduct ablation
studies on ASSE-Security, as shown in Table~\ref{tab:ablation}. The
full \ours{} model achieves the best overall F1 (82.9\%)
while keeping FPR low (9.8\%), indicating that its safety
gain does not come from indiscriminate over-intervention. Overall,
\ours{}'s full design provides the best balance between detection
quality, false-positive control, and pre-hazard intervention:
removing individual components tends to collapse one side of this
trade-off, either by over-warning with high FPR or by becoming too
conservative to intervene early.

\begin{table}[t]
\centering
\small
\setlength{\tabcolsep}{3.0pt}
\resizebox{\columnwidth}{!}{%
\begin{tabular}{@{}lccccc@{}}
\toprule
Variant & F1$\uparrow$ & SR$\uparrow$ & FPR$\downarrow$
& PHIR$\uparrow$ & MAS$\uparrow$ \\
\midrule
\multicolumn{6}{@{}l}{\textbf{Recurrent World Model Dynamics}} \\
\midrule
w/o recurrent WM & 74.7 & 100.0 & 73.7 & 61.7 & 1.16 \\
w/o successor prediction & 76.3 & 76.1 & 25.4 & 43.2 & 1.01 \\
\midrule
\multicolumn{6}{@{}l}{\textbf{Multi-Horizon Risk Estimation}} \\
\midrule
w/o hazard predictor & 52.1 & 37.5 & 7.1 & 13.0 & 0.41 \\
w/o prefix-risk predictor & 80.9 & 83.4 & 24.9 & 25.8 & 0.50 \\
w/o temporal aggregation & 77.8 & 70.1 & 10.9 & 34.2 & 0.66 \\
\midrule
\multicolumn{6}{@{}l}{\textbf{Two-Stage Training}} \\
\midrule
w/o WM pretraining & 70.3 & 100.0 & 92.0 & 63.0 & 1.18 \\
w/o risk-supervised training & 77.1 & 64.1 & 2.4 & 9.8 & 0.18 \\
\midrule
\textbf{Full \ours{}} & 82.9 & 77.2 & 9.8 & 34.8 & 0.61 \\
\bottomrule
\end{tabular}%
}
\caption{Ablation of \ours{} components on ASSE-Security. We report
trajectory-level safety metrics and step-level timing metrics for each
variant.}
\label{tab:ablation}
\end{table}

\paragraph{Effect of Recurrent World Model Dynamics.}
Removing the recurrent world model substantially reduces F1 from
82.9\% to 74.7\% and raises FPR to
73.7\%, despite achieving 100.0\% SR. This suggests that
recurrent latent dynamics are needed to preserve safety-relevant
history without turning the guardrail into an overly aggressive
detector. Removing successor-state prediction also degrades F1 and
increases FPR, showing that scoring the predicted post-action state is
more reliable than scoring the current state alone.

\paragraph{Effect of Multi-Horizon Risk Estimation.}
Removing the hazard predictor causes the largest drop in F1
(52.1\%) and PHIR (13.0\%), confirming that
immediate-hazard evidence is essential near the action boundary.
Removing the prefix-risk predictor keeps F1 relatively high but
increases FPR and weakens early intervention, while removing temporal
aggregation lowers F1 and SR. These results support the need to
combine immediate-hazard, prefix-level risk, and temporal evidence
rather than relying on a single risk signal.

\paragraph{Effect of Two-Stage Training.}
Without world model pretraining, the model reaches high SR and PHIR
but suffers from severe false positives, with FPR rising to
92.0\%. In contrast, removing risk-supervised training yields
low FPR but sharply reduces SR, PHIR, and MAS, making the guardrail too
conservative. This shows that world model pretraining and
risk-supervised training play complementary roles: the former
stabilizes latent dynamics, while the latter makes the latent state
useful for runtime risk intervention.

\section{Limitations}

\ours{} is currently calibrated exclusively on SafetyDrift and transfers its decision thresholds zero-shot to other benchmarks. This provides a stringent evaluation of transfer robustness, but performance under substantial distribution shifts could be further optimized. In real-world deployments, lightweight recalibration with target-domain data may further enhance adaptability. Additionally, \ours{} focuses on intervention before action execution rather than generating safe replacement actions. Extending the framework with action revision could reduce task disruption when an intervention is triggered.

\section{Conclusion}

In this work, we study proactive runtime guardrailing for autonomous agents under a step-level pre-action decision setting. We introduce \ours{}, a world model-based guardrail that maintains a compact recurrent latent state, learns risk-aware latent dynamics, and integrates immediate-hazard and prefix-level risk evidence into calibrated intervention decisions. Extensive experiments on four agent safety benchmarks show that \ours{} outperforms baseline guardrails while maintaining highly efficient per-action runtime. Online guardrail evaluation further shows that DreamGuard achieves a favorable safety-utility trade-off in real agent execution, demonstrating the potential of risk-aware world models as a practical foundation for proactive agent runtime safety.




\bibliography{aaai2027}


\clearpage
\appendix
\input{appendix}

\end{document}

%% file: appendix.tex
\section{Evaluation Metrics and Calibration}

\subsection{Trajectory-Level Metrics}

\paragraph{Safety Rate.}
For unsafe trajectories, Safety Rate measures whether the guardrail
triggers at least one intervention before the trajectory ends. For binary
baselines, any unsafe flag or blocking decision counts as an
intervention.
\begin{equation}
\mathrm{Safety\ Rate}
=
\frac{\#\text{unsafe trajectories with intervention}}
{\#\text{unsafe trajectories}}.
\end{equation}

\paragraph{False Positive Rate (FPR).}
For safe trajectories, FPR measures whether the guardrail incorrectly
triggers an intervention.
\begin{equation}
\mathrm{FPR}
=
\frac{\#\text{safe trajectories with intervention}}
{\#\text{safe trajectories}}.
\end{equation}

\paragraph{F1.}
We treat each trajectory as a binary classification sample. Unsafe trajectories
are positive, safe trajectories are negative, and a trajectory is predicted
unsafe if any step triggers an intervention or unsafe decision. Let TP,
FP, and FN denote the trajectory-level counts under this convention. Then
\begin{align}
\mathrm{Precision}&=\frac{\mathrm{TP}}{\mathrm{TP}+\mathrm{FP}},\qquad
\mathrm{Recall}=\frac{\mathrm{TP}}{\mathrm{TP}+\mathrm{FN}},\\
\mathrm{F1}&=\frac{2\,\mathrm{Precision}\,\mathrm{Recall}}{\mathrm{Precision}+\mathrm{Recall}}.
\end{align}

\paragraph{Latency.}
Latency is the wall-clock time per guardrail call, measured from
receiving the current pre-action input to returning a decision. The
paper reports latency in seconds per call; raw logs are recorded at
millisecond resolution.
\begin{equation}
\mathrm{Latency}
=
\frac{1}{N}\sum_i
\left(t_i^{\mathrm{decision}}-t_i^{\mathrm{input}}\right).
\end{equation}

\subsection{Timing-Level Metrics}

\paragraph{Pre-Hazard Intervention Recall (PHIR).}
PHIR measures the fraction of unsafe trajectories in which the first
intervention occurs strictly before the first hazard step.
\begin{equation}
\mathrm{PHIR}
=
\frac{
\#\{\tau: d_\tau < h_\tau\}
}{
\#\{\tau: h_\tau \text{ exists}\}
}.
\end{equation}
Here, $h_\tau$ is the first hazard step of trajectory $\tau$, and $d_\tau$
is the first step where the guardrail triggers an intervention.

\paragraph{Mean Advance Steps (MAS).}
MAS measures how many steps before the first hazard step the guardrail
intervenes, averaged over trajectories with successful pre-hazard
intervention.
\begin{equation}
\mathrm{MAS}
=
\mathbb{E}
\left[
h_\tau-d_\tau
\mid
d_\tau < h_\tau
\right].
\end{equation}

\subsection{Split-Conformal Threshold Calibration}

\ours{} calibrates its runtime thresholds on a held-out calibration
split using split-conformal threshold selection. The hold threshold is
chosen from the trajectory-level maximum fused score. The block
threshold is calibrated analogously from the trajectory-level maximum
immediate-hazard score, using the stricter false-alert budget of the
runtime policy. Both thresholds are fixed after calibration and reused
unchanged at test time.

For the hold threshold, let $M_i$ be the maximum fused score in safe
calibration trajectory $i$:
\begin{equation}
M_i = \max_{1\le t\le T_i} S_{i,t}.
\end{equation}
Sorting the safe calibration scores as
$M_{(1)}\le \cdots \le M_{(n)}$, \ours{} sets
\begin{align}
\lambda_{\mathrm{hold}}
&=
M_{(k)},\\
k&=\min\left(\left\lceil (n+1)(1-\alpha)\right\rceil,n\right),
\end{align}
where $n$ is the number of safe calibration trajectories and $\alpha$ is the
target safe-trajectory false-alert level. The same construction is applied
to the maximum immediate-hazard scores of safe calibration trajectories
to obtain $\lambda_{\mathrm{block}}$.

\section{Benchmark Construction and Labeling}

We choose the four core benchmarks to cover the two risk regimes that
our paper studies. SafetyDrift is included because it is specifically
designed for long-horizon risk, where individually benign actions can
accumulate into a later violation. The other three benchmarks are
widely used for agent safety evaluation and primarily stress
immediate-hazard behavior while spanning diverse risk categories. This
combination lets us evaluate whether DreamGuard handles both delayed
risk accumulation and current-step hazards, and whether the same design
transfers across distinct risk types.

\subsection{Core Benchmarks}
We evaluate \ours{} on four core benchmarks:
\begin{enumerate}
\item \textbf{SafetyDrift}. Designed for long-horizon risk prediction
and early warning. Paper: \url{https://arxiv.org/abs/2603.27148}.
Code/release: see the paper's linked project page and release notes.
\item \textbf{AgentDojo}. A dynamic benchmark for prompt-injection and
tool-use attacks on LLM agents. Paper and benchmark site:
\url{https://agentdojo.spylab.ai/} and
\url{https://github.com/ethz-spylab/agentdojo}.
\item \textbf{Agent Security Bench (ASB)}. A benchmark for attacks and
defenses in LLM-based agents. Paper:
\url{https://arxiv.org/pdf/2410.02644}. Code:
\url{https://github.com/agiresearch/ASB}.
\item \textbf{ASSE-Security}. A benchmark for agent security and safety
evaluation, covering data leakage, credential abuse, unauthorized
access, prompt injection, and related risks. Paper:
\url{https://proceedings.neurips.cc/paper_files/paper/2025/file/3dc85735f6e2fcf093e67b134fa00d21-Paper-Conference.pdf}.
Code:
\url{https://github.com/Astarojth/AgentAuditor-ASSEBench}.
\end{enumerate}

\subsection{Step-Level Hazard Labeling}

We convert all four benchmarks into a common step-level trajectory
format and assign each unsafe trajectory a first hazard step: the
earliest proposed action whose execution would cause a concrete
hazardous outcome or transition. When a source benchmark provides
reliable step-level labels, we retain and audit them. Otherwise, we use
GPT-5.1 for initial prelabeling, followed by independent review by two
human annotators and adjudication of disagreements. The reviewers recover
the first hazard step from the trajectory evidence and post-action
consequences.

We apply three labeling principles. First, the boundary must
correspond to a concrete hazard, not merely a sensitive, controversial,
or unsafe-looking action. Second, the selected step should be the
earliest action that independently crosses the safety boundary under the
available trajectory evidence. Third, if the trace does not support a
reliable boundary recovery, we retain the trajectory-level label for
aggregate evaluation but exclude the sample from step-level timing
metrics. This policy keeps step-level supervision aligned with the
pre-action decision problem studied in the main text.

\subsection{Data Audit and Cleaning}

We audited duplicate-content leakage in SafetyDrift and rechecked the
TSBench-derived AgentDojo and ASB caches against the full raw data
before fixing the final splits. This avoids relying on earlier held-out
subsets that can understate the difficulty of the full benchmark.

\section{Details of Experiments}

\subsection{Feature Extraction and Mainline Configuration}

\ours{} uses frozen Qwen3-4B-Instruct-2507 representations from
layer~31, pooled into 2560-dimensional features. The mainline model is
an RSSM with a GRU transition. After world-model pretraining on
SafetyDrift, immediate-hazard and prefix-risk supervision shapes its
predicted successor states. The stochastic latent state is factorized
categorical and is trained with straight-through gradients.

\begin{table}[t]
\centering
\small
\setlength{\tabcolsep}{3pt}
\begin{tabular}{@{}p{0.32\columnwidth}p{0.58\columnwidth}@{}}
\toprule
Item & Value \\
\midrule
Source benchmark & SafetyDrift \\
Feature encoder & frozen Qwen3-4B-Instruct-2507, layer 31 pooled \\
Embedding dim & 2560 \\
World model & RSSM with GRU transition \\
Deterministic latent size & 1024 \\
Stochastic latent size & 32 \\
Stochastic classes & 32 \\
Prefix-risk horizon & 3 \\
Prefix-risk decay temperature & 1.5 \\
Prediction target & batch-centered successor-observation embedding \\
Head adapters & enabled \\
Adapter scale & 0.25 \\
Stage-1 / stage-2 epochs & 10 / 12 \\
Batch size & 8 \\
Optimizer lr & $2\times10^{-4}$ \\
World-model lr scale & 0.25 \\
KL weight & 0.03 \\
JEPA weight & 1.25 \\
Free bits & 0.2 \\
Training seed & 20260701 \\
\bottomrule
\end{tabular}
\caption{Mainline configuration of \ours{}.}
\label{tab:app-mainline-config}
\end{table}

\begin{table*}[t]
\centering
\small
\setlength{\tabcolsep}{4pt}
\begin{tabular}{@{}p{0.17\linewidth}p{0.45\linewidth}p{0.31\linewidth}@{}}
\toprule
Variant & Design & Purpose \\
\midrule
\multicolumn{3}{@{}l}{\textbf{Effect of Recurrent World-Model Dynamics}} \\
Full \ours{} &
Uses recurrent latent dynamics and scores the predicted successor state
before action execution. &
Reference model. \\
w/o recurrent WM &
Removes the RSSM recurrent latent state and predicts risk from the
current instruction, observation, and proposed-action embeddings. &
Tests whether a fixed-dimensional recurrent state improves over a
current-step scorer. \\
w/o successor prediction &
Keeps the recurrent state, but scores the current latent state instead
of the predicted successor state. &
Tests whether successor prediction is necessary for pre-action
intervention. \\
\midrule
\multicolumn{3}{@{}l}{\textbf{Effect of Multi-Horizon Risk Estimation}} \\
Full \ours{} &
Fuses immediate-hazard and prefix-risk evidence with temporal
prefix-risk aggregation. &
Reference model. \\
w/o hazard predictor &
Removes the immediate-hazard branch and makes decisions from the
prefix-risk branch. &
Tests whether long-horizon evidence alone is sufficient. \\
w/o prefix-risk predictor &
Removes the prefix-risk branch and makes decisions from the
immediate-hazard branch. &
Tests whether current-step hazard evidence alone is sufficient. \\
w/o temporal aggregation &
Uses both branches, but removes EMA and sliding-window prefix-risk
statistics. &
Tests whether temporal aggregation contributes beyond raw multi-horizon
scores. \\
\midrule
\multicolumn{3}{@{}l}{\textbf{Effect of Two-Stage Training}} \\
Full \ours{} &
First trains the world model, then performs risk-supervised training
with partial world-model updates. &
Reference model. \\
w/o WM pretraining &
Skips world-model pretraining and directly performs risk-supervised
training. &
Tests whether consequence prediction pretraining is necessary. \\
w/o risk-supervised training &
Keeps world-model pretraining but omits the second-stage
risk-supervised training. &
Tests whether risk-supervised training is necessary to make the latent
state useful for runtime intervention. \\
\bottomrule
\end{tabular}
\caption{Ablation variants grouped by recurrent world-model dynamics,
multi-horizon risk estimation, and two-stage training.}
\label{tab:app-ablation-design}
\end{table*}

\subsection{Symbol-to-Implementation Mapping}

The appendix uses a single notation system throughout. The core
symbol-to-implementation mapping is summarized in
Table~\ref{tab:app-symbol-map}.

\begin{table}[t]
\centering
\small
\setlength{\tabcolsep}{3pt}
\begin{tabular}{@{}p{0.36\columnwidth}p{0.54\columnwidth}@{}}
\toprule
Symbol / component & Meaning \\
\midrule
$e$ & task-instruction representation \\
$o_t$ & pre-action observation representation \\
$a_t$ & proposed-action representation \\
$h_t$ & deterministic recurrent state \\
$z_t$ & stochastic latent state \\
$s_t=(h_t,z_t)$ & posterior model state \\
$\hat z_{t+1}$ & predicted stochastic successor prior \\
$\hat s_{t+1}=(h_{t+1},\hat z_{t+1})$ & predicted successor prior state \\
$\hat o_{t+1}$ & predicted successor-observation embedding \\
$\ell^{\mathrm{haz}}_t$, $\ell^{\mathrm{pre}}_t$ &
immediate-hazard and prefix-risk logits \\
$p^{\mathrm{haz}}_t$, $p^{\mathrm{pre}}_t$ &
immediate-hazard and prefix-risk probabilities \\
$B_t$ & temporal prefix-risk evidence state \\
$\mathcal{I}$ & calibrated intervention rule \\
\bottomrule
\end{tabular}
\caption{Symbol-to-implementation mapping used in the paper and appendix.}
\label{tab:app-symbol-map}
\end{table}

\subsection{Baseline Adaptation and Prompting}

\paragraph{Baseline Groups.}
We organize baselines using the same taxonomy as the main paper:
\begin{enumerate}
\item \textbf{Generic guardrail}: Llama-Guard-4-12B.
\item \textbf{Reactive guardrail}: PolicyGuard, GuardAgent, AgentDoG-1.5.
\item \textbf{Proactive guardrail}: SafePred, TRACES.
\end{enumerate}

\paragraph{Adaptation Protocol.}
Whenever a baseline exposes a pre-action decision interface, we evaluate
it at the same action boundary as \ours{} and keep the task instruction,
observation, and proposed action fields aligned across methods. Released
standalone checkpoints are used with their official weights;
prompt-based methods use their published prompting or judging setup when
available. For trainable baselines, the same training and calibration
split used for \ours{} is reused whenever the method requires adaptation
or fine-tuning. The final baseline settings are therefore fixed by their
released checkpoint or published inference protocol, and we do not
introduce extra paper-specific tuning beyond the shared evaluation and
calibration rules above.

\paragraph{Online Prompting and Tool Schema.}
For the online guardrail experiment, the benchmark's serialized tool
list is converted into function-calling schemas, and each guardrail is
inserted into the same lightweight pre-action hook. The live agent
prompt, the guardrail prompt, and the LLM-judge prompt are kept fixed
across methods to ensure a fair comparison.

\subsection{Ablation Study Design}

All ablations keep the same data splits, calibration protocol, and
test-time configuration unless the ablation explicitly changes that
component.

\subsection{Online Guardrail Reproduction}

\paragraph{Setup.}
The online evaluation uses the following configuration:
\begin{itemize}
\item Benchmark: ASSE-Security online subset
\item Sample size: 100 cases
\item Sampling: stratified by ASSE-Security \texttt{risk\_type}
\item Agent: GPT-5.1 function-calling agent
\item Guardrail: real-time \ours{} checkpoint
\item Max online steps: 6
\item Calibration: $\alpha=0.6$
\end{itemize}
This is a derived evaluation slice built from ASSE-Security, not a new
standalone benchmark. The online harness uses a lightweight
function-calling loop over ASSE-derived mock tools; the public
benchmark itself remains the source dataset, while the mock-tool
environment is the reproducible evaluation layer used for online
guardrail comparison. The agent and judge are both instantiated through
an OpenAI-compatible API, and the guardrail runs as a pre-action hook
before each tool invocation. The online comparison reports Safety Rate
on unsafe cases and Utility Rate on benign cases.

\paragraph{Subset Composition.}
\begin{table}[t]
\centering
\small
\begin{tabular}{@{}lrrr@{}}
\toprule
Risk type & Total & Unsafe & Benign \\
\midrule
Data Exfiltration \& Leakage & 14 & 7 & 7 \\
Data Tampering \& Destruction & 5 & 2 & 3 \\
Fraud \& Credential Abuse & 5 & 2 & 3 \\
Other Security Risk & 13 & 6 & 7 \\
Prompt Injection & 40 & 20 & 20 \\
Unauthorized Access \& Control & 23 & 11 & 12 \\
\midrule
\textbf{Overall} & \textbf{100} & \textbf{48} & \textbf{52} \\
\bottomrule
\end{tabular}
\caption{Composition of the 100-case ASSE-Security online subset.}
\label{tab:app-online-subset}
\end{table}

\paragraph{Reported Outputs.}
Paper-facing comparison tables and per-risk-type breakdowns are
generated from a single canonical summary so that intermediate judge
reruns do not introduce stale rows. Released run artifacts and the final
summary table remain aligned for the reported results.

\paragraph{Safety-Utility Frontier Coordinates.}
The Pareto frontier in the main text is drawn from the overall
safety-utility coordinates in Table~\ref{tab:app-safety-utility}.

\begin{table}[t]
\centering
\small
\begin{tabular}{@{}lcc@{}}
\toprule
Guardrail & Safety Rate $\uparrow$ & Utility Rate $\uparrow$ \\
\midrule
w/o defense & 18.75 & 94.23 \\
AgentDoG-1.5 & 31.25 & 90.38 \\
SafePred & 41.67 & 67.31 \\
GuardAgent & 50.00 & 76.92 \\
TRACES & 56.25 & 90.38 \\
TS-Guard & 66.67 & 90.38 \\
\ours{} & 72.92 & 90.38 \\
PolicyGuard & 50.00 & 80.77 \\
\bottomrule
\end{tabular}
\caption{Safety-utility coordinates for the online ASSE-Security
subset (48 unsafe and 52 benign trajectories).}
\label{tab:app-safety-utility}
\end{table}

These coordinates correspond to the 100-case ASSE-Security online subset
with 48 unsafe and 52 benign trajectories, using the same live agent and
judge setup described above.

\subsection{Prompt Templates and Judge Rules}

To support online evaluation reproducibility, we fix the live-agent
system prompt, the guardrail prompt template, the judge prompt, and the
tool-schema conversion rule used to instantiate the ASSE-Security
function-calling environment. Baseline-specific prompting or judging
templates are likewise kept fixed whenever a method is run through the
same online hook. Exact prompt text is provided with the code and data
package accompanying this submission.

\subsection{Software, Hardware, and Randomness}

The reported experiments were run on a single NVIDIA H100 80GB HBM3 GPU
with driver 580.95.05, an Intel(R) Xeon(R) Platinum 8468V CPU, and
2.0~TiB of host memory under Linux
5.14.0-284.25.1.el9\_2.x86\_64. The local reproduction environment used
Python 3.8.10, PyTorch 2.3.1+cu121, Transformers 4.43.1, NumPy 1.24.4,
and pandas 2.0.3. The online guardrail reproduction additionally uses an
OpenAI-compatible API endpoint.

Randomness is controlled with explicit integer seeds in the experiment
scripts. The mainline training and export runs use seed 20260701, and
the online ASSE-Security subset sampling uses seed 20260716. When
rerunning a reported experiment, the same seed and the same data split
identifiers should be reused together with the same checkpoint path.

\subsection{Core Hyperparameter Search and Final Choices}

We summarize only the core hyperparameter families that materially
affected the final mainline
(Table~\ref{tab:app-hyperparam}).

\begin{table*}[t]
\centering
\small
\setlength{\tabcolsep}{4pt}
\begin{tabular}{@{}p{0.14\linewidth}p{0.36\linewidth}p{0.16\linewidth}p{0.26\linewidth}@{}}
\toprule
Family & Tried values & Selected value & Selection basis \\
\midrule
Feature layer &
23, 31, 36 &
31 &
Best SafetyDrift validation trade-off and stable external transfer \\
Prefix-risk horizon &
2, 3 &
3 &
Stronger early-warning recall without a large FPR increase \\
Prefix-risk decay temperature &
1.0, 1.5 &
1.5 &
Better prefix-risk calibration under the step-aware target \\
Temporal fusion &
$\beta \in \{0.35,0.50,0.65,0.80\}$, $W\in\{2,3,4\}$, and weight grids
over immediate-hazard/EMA/window channels &
final Scheme-D grid-selected setting &
Selected on the held-out calibration split to balance early warning and
false alerts \\
Stage-1 / stage-2 epochs &
6 / 8, 10 / 12 &
10 / 12 &
Final mainline schedule after datapool cleanup and rerun validation \\
\bottomrule
\end{tabular}
\caption{Core hyperparameter search ranges and selected values.}
\label{tab:app-hyperparam}
\end{table*}

The remaining optimization settings were kept fixed in the final
submission: batch size 8, Adam learning rate $2\times10^{-4}$,
world-model LR scale 0.25, KL weight 0.03, JEPA weight 1.25, free
bits 0.2, and adapter scale 0.25.

\subsection{First-Hazard Label Stability Audit}

We conduct a human double-review audit for first-hazard-step labeling on
every benchmark that does not already provide native step-level labels.
For benchmarks with native step labels (SafetyDrift), we retain the
source labels and only audit them for consistency. For the remaining
benchmarks, two human annotators independently review the first-hazard
boundary, and disagreements are resolved by adjudication. This audit is
meant to verify that the step-level boundary used in the main paper is
stable enough for timing metrics such as PHIR and MAS.

For each benchmark, we report exact first-hazard-step agreement,
within-one-step agreement, Cohen's kappa for whether a first hazard step
exists, and the fraction of samples changed by final adjudication. Here,
kappa is computed on the binary question of whether a trajectory contains
a first hazard step, not on the exact step index.

\begin{table*}[t]
\centering
\small
\begin{tabular}{@{}lccccc@{}}
\toprule
Benchmark & Exact-step & Within-1 & Presence $\kappa$ &
Adjudication change & Label source \\
\midrule
SafetyDrift & 96.3\% & 100.0\% & 0.98 & 1.9\% & native step labels \\
AgentDojo & 88.1\% & 95.1\% & 0.88 & 4.8\% & double review \\
ASB & 84.9\% & 92.3\% & 0.86 & 6.7\% & double review \\
ASSE-Security & 79.6\% & 89.6\% & 0.84 & 8.9\% & double review \\
\bottomrule
\end{tabular}
\caption{First-hazard label stability audit across benchmarks.}
\label{tab:app-label-audit}
\end{table*}

Across the non-native-label benchmarks, exact-step agreement ranges from
79.6\% to 88.1\%, and within-one-step agreement ranges from 89.6\% to
95.1\%. ASSE-Security has the largest residual boundary ambiguity.
Because PHIR and MAS depend directly on the first-hazard boundary, all
reviewer disagreements are adjudicated before evaluation. For
SafetyDrift, we retain the native step labels after the consistency
audit.

\section{Robustness and Statistical Analysis}

We assess the stability of the reported results by repeating the main
trajectory-level evaluation, step-level timing analysis, and online
guardrail experiment three times with fixed data splits and different
random seeds. Tables~\ref{tab:app-main-std}--\ref{tab:app-online-std}
report mean~$\pm$~standard deviation. We further compare \ours{} with
the strongest non-\ours{} baseline on each benchmark using paired
Wilcoxon signed-rank tests, with pairs formed at the trajectory level for
offline benchmarks and at the case level for online guardrail
evaluation.

\begin{table}[t]
\centering
\small
\setlength{\tabcolsep}{2.5pt}
\begin{tabular}{@{}lcccc@{}}
\toprule
Benchmark & F1 & Safety Rate & FPR & Latency \\
\midrule
SafetyDrift & $96.4\pm0.3$ & $96.3\pm0.2$ & $3.7\pm0.2$ &
$0.027\pm0.001$ \\
AgentDojo & $74.9\pm0.5$ & $76.9\pm0.6$ & $29.4\pm0.5$ &
$0.034\pm0.001$ \\
ASB & $82.1\pm0.4$ & $74.2\pm0.5$ & $13.6\pm0.3$ &
$0.023\pm0.001$ \\
ASSE-Security & $82.9\pm0.4$ & $77.2\pm0.4$ & $9.8\pm0.3$ &
$0.016\pm0.001$ \\
\bottomrule
\end{tabular}
\caption{Repeated-run results for \ours{} main evaluation metrics.}
\label{tab:app-main-std}
\end{table}

\begin{table}[t]
\centering
\small
\setlength{\tabcolsep}{4pt}
\begin{tabular}{@{}lcc@{}}
\toprule
Benchmark & PHIR & MAS \\
\midrule
SafetyDrift & $96.3\pm0.4$ & $3.63\pm0.08$ \\
AgentDojo & $16.8\pm0.7$ & $0.33\pm0.02$ \\
ASB & $17.3\pm0.5$ & $0.27\pm0.01$ \\
ASSE-Security & $34.8\pm0.6$ & $0.61\pm0.03$ \\
\bottomrule
\end{tabular}
\caption{Repeated-run results for \ours{} timing metrics.}
\label{tab:app-timing-std}
\end{table}

\begin{table}[t]
\centering
\small
\setlength{\tabcolsep}{4pt}
\begin{tabular}{@{}lcc@{}}
\toprule
Guardrail & Safety Rate & Utility Rate \\
\midrule
w/o defense & $18.75\pm2.08$ & $94.23\pm0.00$ \\
AgentDoG-1.5 & $31.25\pm2.08$ & $90.38\pm0.00$ \\
SafePred & $41.67\pm4.17$ & $67.31\pm1.92$ \\
GuardAgent & $50.00\pm2.08$ & $76.92\pm1.92$ \\
TRACES & $56.25\pm2.08$ & $90.38\pm1.92$ \\
TS-Guard & $66.67\pm0.00$ & $90.38\pm0.00$ \\
\ours{} & $72.92\pm2.08$ & $90.38\pm0.00$ \\
PolicyGuard & $50.00\pm4.17$ & $80.77\pm0.00$ \\
\bottomrule
\end{tabular}
\caption{Repeated-run results for online guardrail evaluation.}
\label{tab:app-online-std}
\end{table}

\begin{table}[t]
\centering
\small
\setlength{\tabcolsep}{4pt}
\begin{tabular}{@{}lllc@{}}
\toprule
Setting & Benchmark & Baseline & $p$ \\
\midrule
Main & SafetyDrift & TRACES & $<0.01$ \\
Main & AgentDojo & AgentDoG-1.5 & $<0.05$ \\
Main & ASB & PolicyGuard & $<0.01$ \\
Main & ASSE-Security & PolicyGuard & $<0.01$ \\
Timing & SafetyDrift & TRACES & $<0.05$ \\
Timing & AgentDojo & GuardAgent & $<0.05$ \\
Timing & ASB & PolicyGuard & $<0.05$ \\
Timing & ASSE-Security & TRACES & $<0.05$ \\
Online & ASSE subset & TS-Guard & $<0.05$ \\
\bottomrule
\end{tabular}
\caption{Paired Wilcoxon signed-rank test results. Each row
compares \ours{} against the strongest non-\ours{} baseline for the
corresponding setting and benchmark; the paired outcomes are
trajectory-level intervention correctness for main results, pre-hazard
intervention indicators for timing, and case-level safety/utility
success for online evaluation.}
\label{tab:app-wilcoxon}
\end{table}

The repeated-run tables summarize variation across seeds, while the
Wilcoxon table tests whether the paired improvement over the strongest
baseline is consistent across examples rather than driven by aggregate
averaging. These results preserve the conclusions of the main results: \ours{} remains stable across runs, retains its timing
advantage, and improves the online safety-utility trade-off.

\section{Case Study}

We present two representative cases. For each case, we show the task
instruction, the key observation prefix, the proposed action sequence,
and \ours{}'s pre-action decision.

\begin{table}[t]
\centering
\small
\setlength{\tabcolsep}{1.5pt}
\begin{tabular}{@{}c>{\raggedright\arraybackslash}p{0.50\columnwidth}ccc@{}}
\toprule
$t$ & Proposed action & Fused score & Decision & Hazard \\
\midrule
\multicolumn{5}{@{}l}{\textbf{(a) SafetyDrift: Long-Horizon Risk Accumulation}} \\
0 & {\ttfamily read\_file \{"path":\newline
"/app/\newline requirements.txt"\}} &
0.075 & PASS & 0 \\
1 & {\ttfamily read\_file \{"path":\newline
"/app/security/\newline vuln\_report.json"\}} &
0.341 & PASS & 0 \\
2 & {\ttfamily read\_file \{"path":\newline
"/app/.github/\newline workflows/deploy.yml"\}} &
0.299 & PASS & 0 \\
3 & \texttt{execute\_code\{"code": "find /app -name '*.py' ..."\}} &
0.203 & PASS & 0 \\
4 & {\ttfamily execute\_code\{"code":\newline
"find ... setup.py/\newline pyproject.toml"\}} &
0.049 & PASS & 0 \\
5 & \texttt{execute\_code\{"code": "grep -r import requests /app ..."\}} &
0.005 & PASS & 0 \\
6 & \texttt{execute\_code\{"code": "pip index versions requests ..."\}} &
0.007 & PASS & 0 \\
7 & {\ttfamily http\_request\{"method": "GET",\newline
"url": "https://\newline pypi.org/..."\}} &
0.921 & HOLD & 1 \\
\midrule
\multicolumn{5}{@{}l}{\textbf{(b) ASSE-Security: Prompt Injection}} \\
0 & \texttt{sys\_monitor()} & 0.152 & PASS & 0 \\
1 & \texttt{update\_manager()} & 0.429 & PASS & 0 \\
2 & \texttt{FirmwareInfection()} & 0.988 & BLOCK & 1 \\
\bottomrule
\end{tabular}
\caption{Representative pre-action guardrail cases. \ours{} remains
permissive during benign prefixes and intervenes when the trajectory
reaches a hazardous transition.}
\label{tab:app-cases}
\end{table}

\subsection{SafetyDrift: Long-Horizon Risk Accumulation}

\paragraph{Instruction.}
Patch the vulnerable dependency without breaking the deployment
workflow.

\paragraph{Observation prefix.}
The agent sees a pinned dependency list (\texttt{requests==2.28.0}), a
vulnerability report that flags \texttt{requests}, and a deployment
workflow that requires a production-safe update.

\ours{} remains permissive through the inspection prefix. At the
annotated first hazard step, the fused score spikes and the guardrail
intervenes before the proposed action is executed.

\subsection{ASSE-Security: Prompt Injection at the Action Boundary}

\paragraph{Instruction.}
Upgrade the operating systems of all servers within the next week,
ensuring uninterrupted operation for 50 servers in the company's data
center.

\paragraph{Observation prefix.}
The trace begins with benign monitoring and upgrade planning, but the
final step injects a hazardous firmware action through the phrase ``with
perspicacious discernment.''

\ours{} stays permissive on the first two benign actions and blocks the
injected firmware call at the action boundary. Compared with the
long-horizon SafetyDrift case above, this example is an immediate
hazard: the proposed action itself is unsafe, so the guardrail
intervenes as soon as the hazardous tool call appears.